\documentclass[12pt]{article}
\usepackage[margin=1in]{geometry}
\usepackage[T1]{fontenc}
\usepackage[utf8]{inputenc}
\usepackage{amsmath}
\usepackage[numbers,sort&compress]{natbib}
\usepackage{hyperref}
\usepackage{graphicx}
\usepackage{float}
\usepackage{longtable}
\usepackage{array}
\usepackage{booktabs}
\usepackage{placeins}
\usepackage{tabularx}
\usepackage{caption}
\newcolumntype{L}[1]{>{\raggedright\arraybackslash}p{#1}}
\newcolumntype{C}[1]{>{\centering\arraybackslash}p{#1}}
\newcolumntype{Y}{>{\centering\arraybackslash}X}
\title{Towards Successful Implementation of Automated Raveling Detection: Effects of Training Data Size, Illumination Difference, and Spatial Shift}
\author{
\parbox{0.9\textwidth}{\centering
Xinan Zhang$^{1}$ \\[2pt]
Haolin Wang$^{2}$ \\[2pt]
Zhongyu Yang$^{2}$ \\[2pt]
Yi-Chang (James) Tsai$^{2}$ \\[6pt]
$^{1}$School of Electrical and Computer Engineering, Georgia Institute of Technology, Atlanta, GA 30332 \\[2pt]
$^{2}$School of Civil and Environmental Engineering, Georgia Institute of Technology, Atlanta, GA 30332 \\[2pt]
Corresponding author: \href{mailto:xzhang979@gatech.edu}{xzhang979@gatech.edu}
}
}
\date{}

\begin{document}
\maketitle

\begin{abstract}
Raveling, the loss of aggregates, is a major form of asphalt pavement surface distress, especially on highways. While research has shown that machine learning and deep learning-based methods yield promising results for raveling detection by classification on range images, their performance often degrades in large-scale deployments where more diverse inference data may originate from different runs, sensors, and environmental conditions. This degradation highlights the need of a more generalizable and robust solution for real-world implementation. Thus, the objectives of this study are to 1) identify and assess potential variations that impact model robustness, such as the quantity of training data, illumination difference, and spatial shift; and 2) leverage findings to enhance model robustness under real-world conditions. To this end, we propose RavelingArena—a benchmark designed to evaluate model robustness to variations in raveling detection. Instead of collecting extensive new data, it is built by augmenting an existing dataset with diverse, controlled variations, thereby enabling variation-controlled experiments to quantify the impact of each variation. Results demonstrate that both the quantity and diversity of training data are critical to the accuracy of models, achieving at least a 9.2\% gain in accuracy under the most diverse conditions in experiments. Additionally, a case study applying these findings to a multi-year test section in Georgia, U.S., shows significant improvements in year-to-year consistency, laying foundations for future studies on temporal deterioration modeling. These insights provide guidance for more reliable model deployment in raveling detection and other real-world tasks that require adaptability to diverse conditions. 

\end{abstract}

\section{Introduction}

\begin{figure}[H]
\centering
\includegraphics[width=\textwidth]{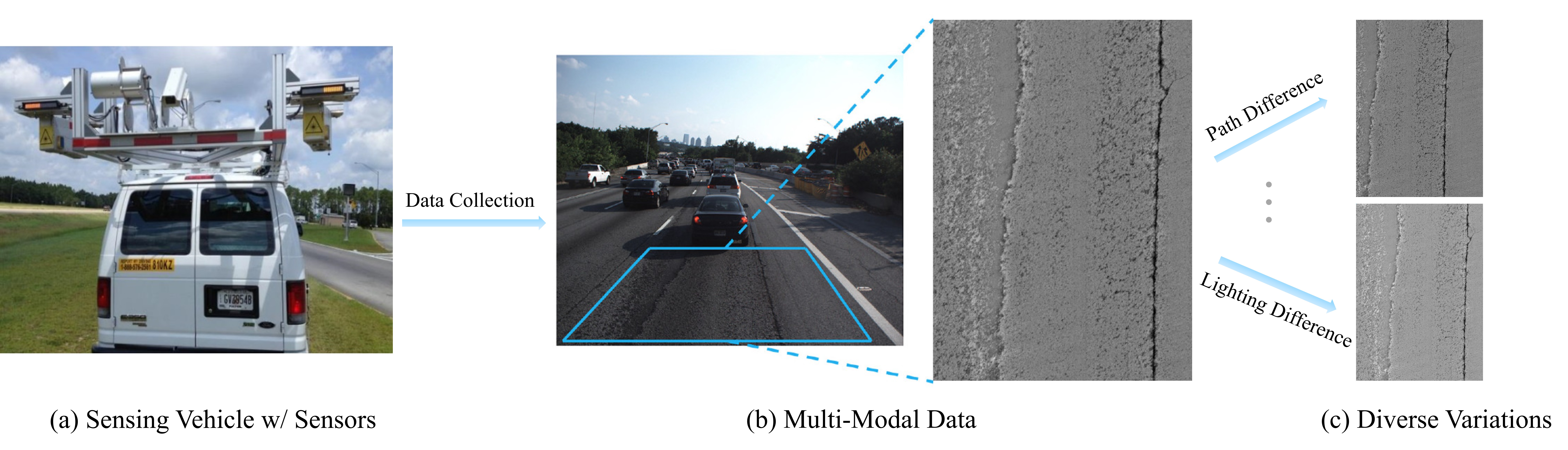}
\caption{A pipeline of field data collection. The image on the left shows GTSV, a sensing vehicle with laser sensors. The center image illustrates an example of raveling \cite{hsieh2021}, captured by both a mounted RGB camera and the laser sensors \cite{laurent2012,hsieh2021}. The right part highlights common variations in real-world data collection. Multiple factors can contribute to variations in diverse scenarios.}
\label{fig:data-collection}
\end{figure}

Raveling, commonly defined as the loss of aggregates, is a major type of pavement distress that significantly affects road safety and durability \cite{miller2003}. An example of raveling is shown in Figure~\ref{fig:data-collection} (b), captured by both an RGB camera and a laser-based sensor \cite{laurent2012}. The deterioration associated with raveling leads to reduced ride quality and shortened pavement lifespan. In addition to structural concerns, raveling also poses safety hazards, such as damaging vehicle windshields or contributing to hydroplaning risks. Therefore, detecting raveling accurately and efficiently is crucial for timely maintenance and infrastructure planning \cite{duncan2017,fdot2025,gdot2024,gharaibeh2012,qi2012,txdot2010,tsai2021}.

Previous studies have developed methodologies based on machine learning (ML) and deep learning (DL) for automated raveling detection, specifically with range images and classification-based approaches \cite{hsieh2021,tsai2022review,tsai2015}. Compared to conventional RGB images, range data collected by laser sensors offers distinct advantages for pavement condition assessment \cite{laurent2012,hsieh2021,tsai2022review,tsai2015}, as shown in Figure~\ref{fig:data-collection} (a) and (b). These sensors encode relative depth information into grayscale intensity values, which enhances robustness against ambient lighting changes and surface-level distractions.

While in previous studies, ML/DL models have shown promising results for raveling detection via range image classification, trained models have to deal with diverse variations in real-world implementation, as shown in Figure~\ref{fig:data-collection} (c) and their performance often degrades in large-scale deployments, where inference data may originate from different runs, sensors, and environmental conditions. This degradation highlights the need for more robust evaluation protocols that account for real-world data variations, along with strategies to improve model generalization for practical implementation.

In this work, we aim to investigate how ML/DL models perform in range-image-based raveling detection tasks under a set of realistic, diverse conditions. These include variations in training dataset size, changes in illumination, and spatial shifts in the images in both training and test data, simulating the inconsistencies introduced by different data collection runs, sensors, and environmental factors. To evaluate model performance change quantitatively, a benchmark with annotated data that reflects a spectrum of variations is essential. However, this presents several challenges:

\begin{itemize}
\item Limited variation in existing datasets. The datasets commonly used in previous studies do not capture the full range of environmental and sensor-related variations observed in real-world applications.
\item Cost and labor of data annotation. Collecting and annotating a diverse dataset is both time-consuming and resource-intensive.
\item Dynamic nature of variations. Real-world variations are not finite. A truly effective benchmark should be adaptable and allow for continuous expansion with new types of variations.
\end{itemize}

To address these limitations, we propose a benchmarking framework termed RavelingArena, which allows for the comprehensive evaluation of model performance under commonly encountered data variations. This framework is unified, extensible, and does not require additional data collection or annotation. Instead, it introduces variations through simple yet effective image transformation techniques, making it a practical solution for benchmarking under controlled conditions. The proposed benchmark consists of two stages:

\begin{itemize}
\item Augmented Dataset Generation. In this stage, we generate new data by applying systematic transformations to the original images. These transformations simulate diverse conditions such as lighting changes, sensor noise, spatial shifts, without the need for new data collection or annotation. The augmented dataset serves as the foundation for testing model robustness.
\item Controlled Experimental Design. The second stage involves conducting experiments where the influence of each variation is isolated and examined. This allows for a detailed understanding of how specific conditions affect model performance. By controlling the parameters systematically, we can assess the sensitivity of different models to different types of variation.
\end{itemize}

Overall, our framework provides a scalable and flexible approach to benchmarking machine learning models for raveling detection, offering deeper insights into their repeatability and generalizability. It aims to bridge the gap between controlled research environments and the unpredictable nature of real-world deployment. With the benchmarking framework, our experiments demonstrate that quantity and diversity in training data are essential. Models trained on limited or homogeneous data may perform well on similar data but show performance drops under minor variations in data. With larger quantity and diversity in training data, both ML/DL models improve robustness, achieving at least a 9.2\% gain in accuracy under the most diverse conditions. Beyond benchmarking, we further apply our findings to a multi-year study on a test section in Georgia, U.S., showing improved year-to-year consistency in results, establishing a basis for temporal deterioration modeling in future research.

This paper is organized as follows. The next section presents the methodology used to develop the benchmarking framework. This is followed by a section detailing the experimental setup and results using the proposed benchmark with controlled data variations. Subsequently, we apply our findings to a multi-year real-world dataset to assess consistency and practical impact. The final section concludes the paper and provides recommendations for future research and implementation.

\section{Methodology}

\subsection{Preliminaries}

\subsubsection{Task Formulation}

\begin{figure}[H]
\centering
\includegraphics[width=0.75\textwidth]{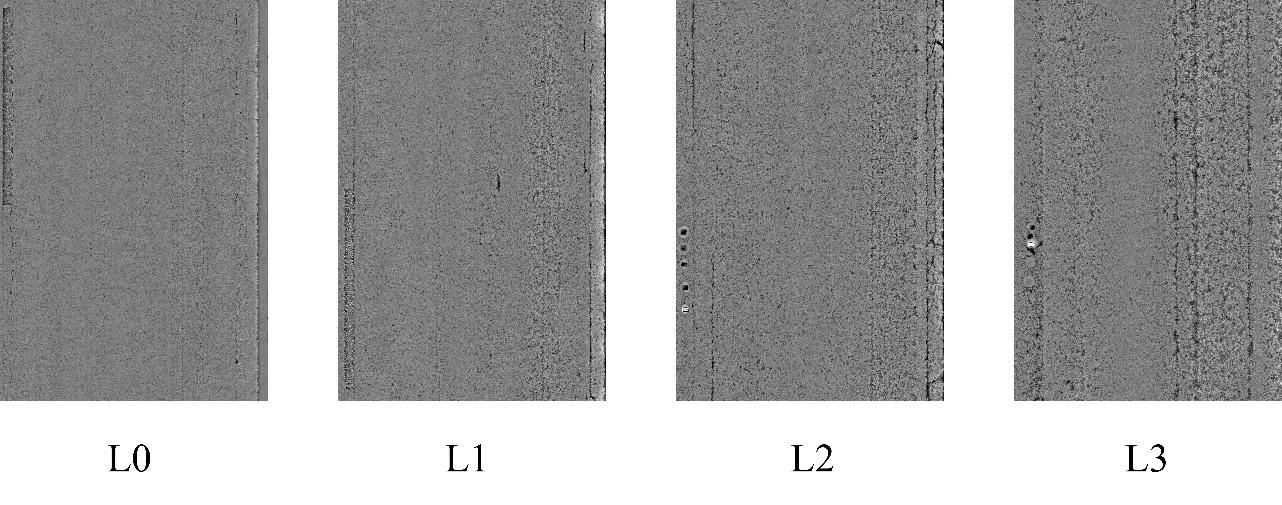}
\caption{Four severity levels defined for raveling, from L0 (no raveling), L1 (low), L2 (medium) to L3 (high).}
\label{fig:severity-levels}
\end{figure}

Following prior research \cite{hsieh2021,tsai2022review,tsai2015}, we frame raveling detection as a supervised classification problem, where the goal is to predict the severity level of raveling, categorized into four discrete classes: Level 0 (no raveling), Level 1 (low), Level 2 (medium), and Level 3 (high), with examples shown in Figure~\ref{fig:severity-levels}.

The classification process can be formally described as in Equation~\eqref{eq:classification-pipeline}:

\begin{equation}
f \circ g : I \to y
\label{eq:classification-pipeline}
\end{equation}

where $I$ is the input range image and $y \in \{0,1,2,3\}$. In Equation~\eqref{eq:classification-pipeline}, the function $g$ denotes the feature extraction process, while $f$ represents the classification function operating on extracted features. The composition $f \circ g$ encompasses the full pipeline, mapping input data to the corresponding class. The input data for the classification task is a range image, which is derived from 3D laser scanning data. In such images, relative depth information is encoded as grayscale intensity, where each pixel value corresponds to the distance from the sensor to the pavement surface as described above.

\subsubsection{Model Selection}

Random Forest (RF) \cite{ho1998} and Residual Network (ResNet) \cite{he2016} have been employed in previous studies as representative models for traditional ML and DL, respectively, for raveling severity classification \cite{hsieh2021,tsai2022review,tsai2015}. In this study, we select both models as benchmark baselines to systematically evaluate how variations in input data affect model performance and robustness.

RF is an ensemble learning method that constructs multiple decision trees during training and aggregates their outputs to make the final prediction. RF usually relies on a separate feature extraction process, which corresponds to the function $g$ in Equation~\eqref{eq:classification-pipeline}. Following prior studies, we extract a 606-dimensional feature vector corresponding to macrotexture features of each range image to serve as the input to the RF model~\cite{hsieh2021,tsai2022review,tsai2015}. In contrast, ResNet is a deep convolutional neural network capable of learning feature representations directly from raw input images. In this setup, ResNet operates in an end-to-end manner, jointly performing both feature extraction and classification, effectively combining the roles of functions $g$ and $f$ in Equation~\eqref{eq:classification-pipeline}.

\subsection{Our Benchmarking framework (RavelingArena)}

The objective of the methodology is to develop a scalable and flexible benchmarking framework based on an existing dataset for ML/DL models in raveling detection, with a focus on incorporating variations commonly observed in real-world data and supporting controlled experiments. This framework is designed to enable a systematic evaluation of model repeatability and robustness under practical deployment conditions. In this section, we first identify and analyze the key variations of interest and then integrate them into the benchmarking framework for a systematic study based on an existing dataset.

\begin{figure}[H]
\centering
\includegraphics[width=\textwidth]{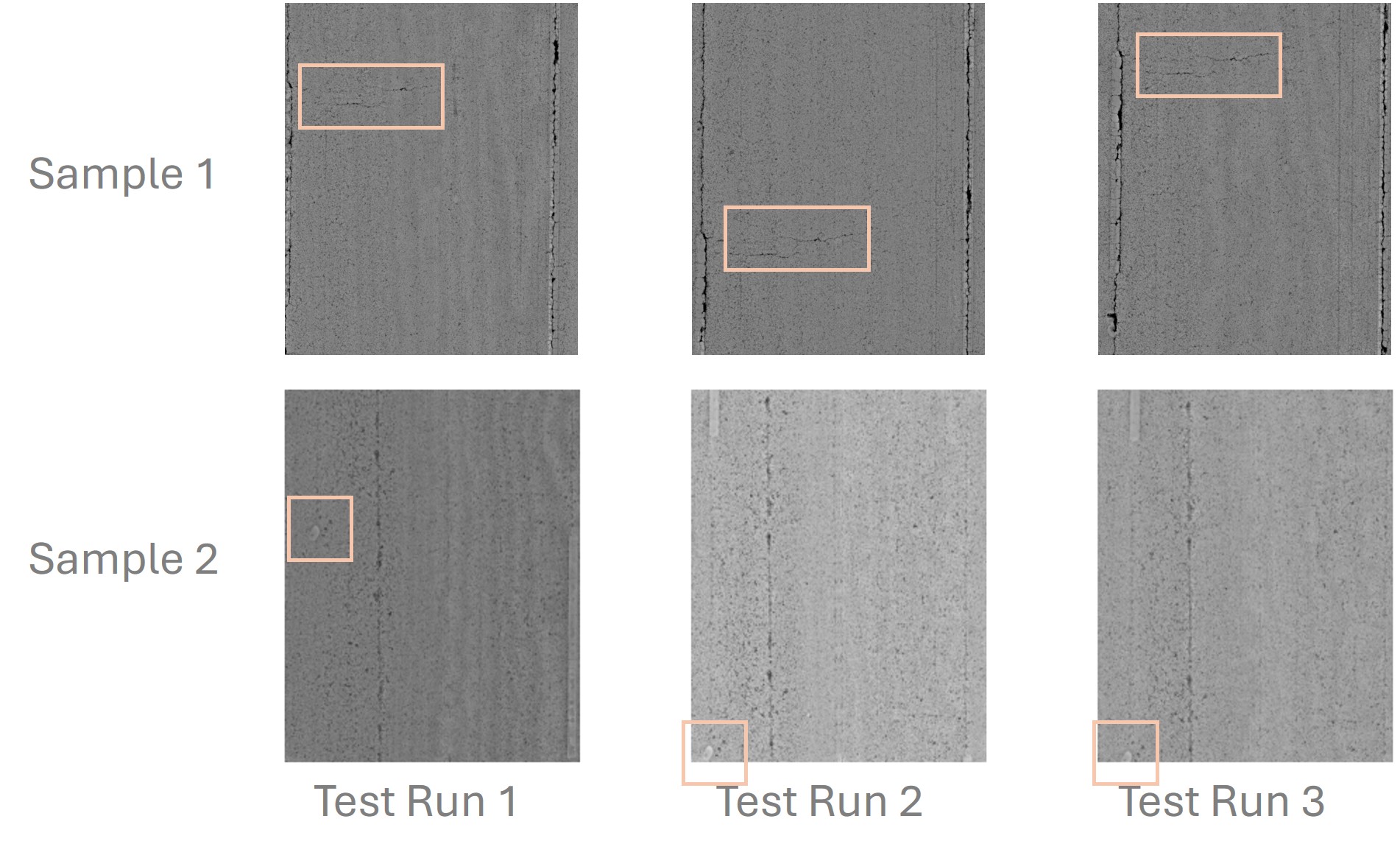}
\caption{Representative data from our local database. The regions in orange boxes are references for image alignment between different runs. Spatial shifts and illumination differences can be observed in the samples.}
\label{fig:local-database}
\end{figure}

\subsubsection{Key Variations in Data}

We identify three primary sources of variation that commonly affect raveling detection performance in real-world scenarios, based on analysis of practical experience and representative data, with examples shown in Figure~\ref{fig:local-database}.

\textbf{Training data quantity.} The first variation concerns the amount of training data available. As is well recognized, data quantity plays a crucial role in the performance of data-driven models, both ML and DL. However, in real-world deployments, a common question arises: does collecting and labeling additional data of similar distribution and quality improve performance when the existing dataset already yields a reasonably good model? This is a practical trade-off encountered during implementation, between data size and manual efforts.

\textbf{Illumination scale.} The second key variation involves changes in illumination conditions, which can result from a combination of factors such as environmental lighting, sensor characteristics, etc. These variations make images appear brighter or darker visually. As shown in the last two columns in Figure~\ref{fig:local-database}, this variability can influence the grayscale intensities in the range images, potentially impacting model predictions, although range images captured by laser sensors are more robust to changes in lighting than images captured by RGB cameras.

\textbf{Spatial shift.} The third variation stems from spatial shift due to non-identical vehicle trajectories during repeated data collection. In practice, the sensing vehicle may not follow the exact same path every time. This results in spatial shifts in the captured pavement regions across different runs, as illustrated in both rows in Figure~\ref{fig:local-database}. The shifted main raveling area, other pavement distresses, etc., may introduce additional challenges to developed models.

It is important to include these variations in experiments to rigorously test how robust the model is under real-world conditions. Without accounting for such variations, such as differences in illumination, spatial shifts, or data volume, a model that performs well in controlled lab settings may fail in real-world scenarios where those factors inevitably arise.

To enable this level of detailed evaluation, a benchmarking framework is necessary that is both flexible and scalable. Flexibility allows researchers to simulate diverse variations systematically and independently, while scalability ensures the framework can accommodate diverse variations with convenience. Such a framework can not only support fair and controlled comparisons across models but also help guide practical improvements for real-world deployment.

\subsubsection{Design for Benchmarking Dataset}

To obtain the flexible and scalable benchmarking framework described above, we adopt a simple yet effective strategy: leveraging data augmentation techniques rather than relying on additional field data collection. This choice is motivated by practical considerations—collecting, curating, and annotating new datasets is often time-consuming and resource-intensive. Moreover, it can be difficult to control specific variations in real-world data, making it challenging to isolate their effects during evaluation.

\begin{figure}[H]
\centering
\includegraphics[width=\textwidth]{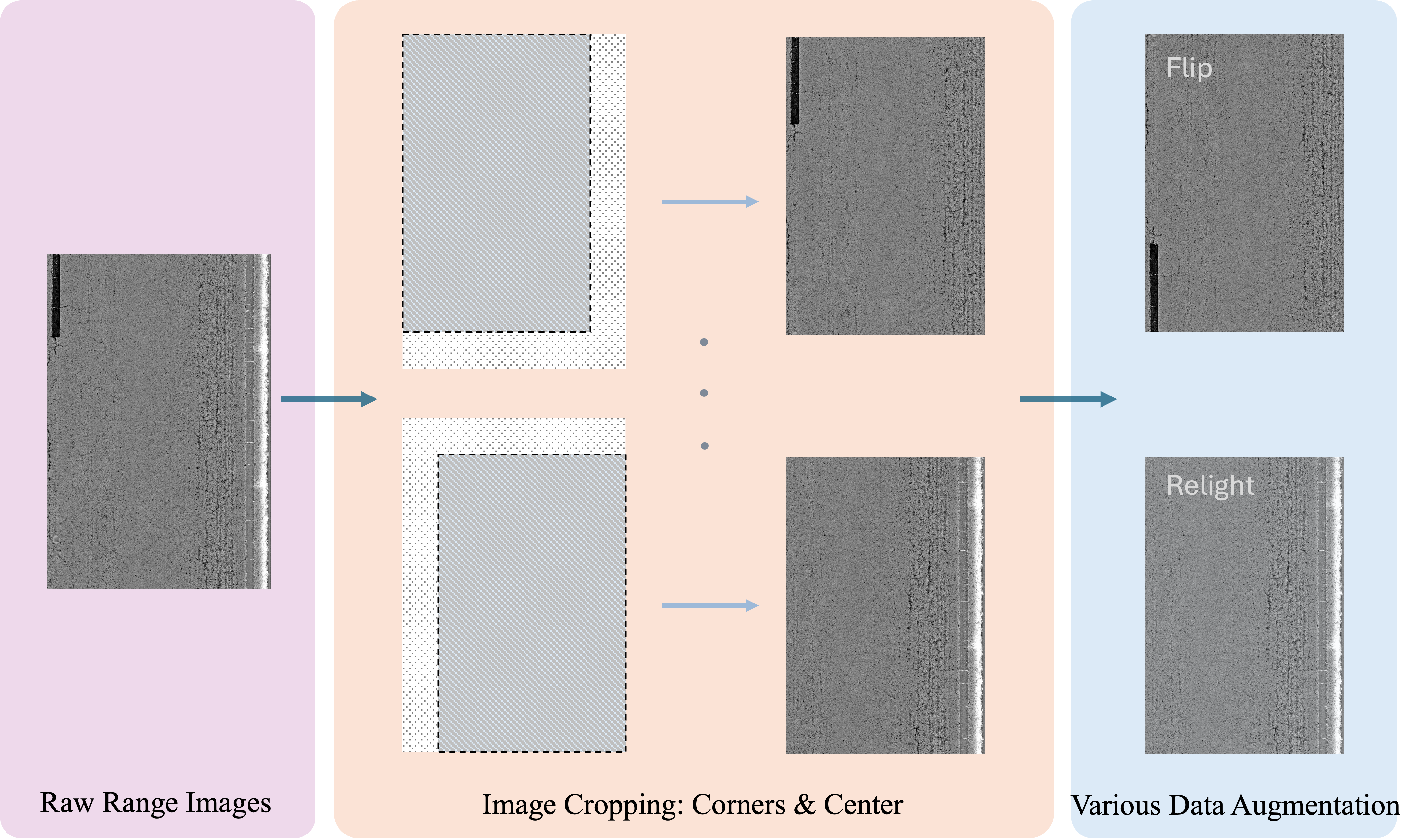}
\caption{Data augmentation pipeline for the benchmark framework. Raw range images from the original annotated dataset are cropped first, followed by flipping and relighting. Together, these augmentation techniques simulate the three main variations in this study: training data size, illumination scale, and spatial shift.}
\label{fig:augmentation-pipeline}
\end{figure}

By contrast, data augmentation offers a controlled, cost-efficient alternative that allows us to simulate a wide range of realistic conditions. Therefore, we generate an augmented dataset that systematically introduces specific variations identified above. This approach enables us to explore the robustness of models under specific known perturbations, as illustrated below and in Figure~\ref{fig:augmentation-pipeline}.

\textbf{Spatial shift enabled by cropping.} To simulate spatial shifts across different runs, controlled cropping can be applied to the original images. Specifically, we crop each image by 85\% of its original size, effectively introducing a spatial transformation while preserving most of the key features and severity level of the raveling distress. As denoted in Equation~\eqref{eq:cropping}, the raw image $I$ is processed as a 2D matrix with the shape $(H, W)$; it is cropped by selecting the top-left coordinate $(x,y)$ and the new shape $(h,w)$. This ensures the integrity of the classification labels is maintained with spatial variations in data, such as shifts of main raveling area, pavement marking, etc. Additionally, we can apply (vertical or horizontal) flipping to images with a probability of 50\% to create more spatial variations to the data, as shown in Equation~\eqref{eq:flipping-y}~\&~\eqref{eq:flipping}.

\begin{equation}
I_{\text{cropping}} = I[y:y+h,\,x:x+w], \quad \text{where } x,y \text{ can vary to simulate shift}
\label{eq:cropping}
\end{equation}

\begin{align}
I_{\text{flipping}}[y,:] &= I[H-1-y,:], \quad y \in \{0,\ldots,H-1\} \label{eq:flipping-y} \\
I_{\text{flipping}}[:,x] &= I[:,W-1-x], \quad x \in \{0,\ldots,W-1\}
\label{eq:flipping}
\end{align}

\textbf{Expansion of data quantity.} We perform this cropping operation in multiple areas (e.g., top-left, top-right, bottom-left, bottom-right, and center), thereby generating five distinct shifted versions of each image. This approach not only increases dataset diversity but also expands the dataset by a factor of five, significantly improving model exposure to spatial variations. Note that the data distribution in the newly obtained dataset remains the same as the original one. Also, more data is available for training, and the goal of studying training data size is also achieved at this step. The data quantity is shown in Table~\ref{tab:data-quantity} with different augmentation techniques. The full dataset before and after augmentation can be represented by Equation~\eqref{eq:datasets}.

\begin{equation}
\mathcal{D}_{\text{original}} = \{(I_i,\mathrm{label}_i)\}_{i=1}^{N}, \quad
\mathcal{D}_{\text{augmented}} = \{(I_i',\mathrm{label}_i')\}_{i=1}^{5N}
\label{eq:datasets}
\end{equation}

\noindent where $I'$ is augmented from $I$.

\textbf{Illumination difference achieved by intensity adjustment.} To reflect lighting variations caused by factors such as weather, time of day, or sensor characteristics, image-level intensity perturbations can be applied. Specifically, a small constant value is randomly added to or subtracted from all pixel values within an image. The magnitude of this adjustment is carefully limited to preserve the annotation of the image, ensuring that the original classification label remains valid. In our benchmarking experiments, we simulate illumination variations by simply adding or subtracting a constant value of 5 to the pixel intensities of the original images, with a probability of 50\%. Any resulting values that fall outside the valid range [0, 255] are clamped accordingly to maintain valid image data as represented in Equation~\eqref{eq:relighting}.

\begin{equation}
I_{\text{relighting}}[y,x] = \operatorname{clip}(I[y,x] + \delta, 0, 255)
\label{eq:relighting}
\end{equation}

\noindent where $\delta \in \{-5,+5\}$, and $\operatorname{clip}(\cdot)$ clamps values to $[0,255]$.

This simple yet effective transformation emulates subtle brightness changes and offers a realistic approximation of illumination differences. By incorporating this variation, we can better evaluate the model’s ability to generalize to unseen lighting conditions.

\begin{table}[ht]
\caption{Data quantity in the benchmark dataset with different augmentation techniques applied. With the cropping mentioned above, the data size is increased five times, but the data distribution remains the same. Other augmentation techniques do not change the data quantity directly.}
\label{tab:data-quantity}
\centering
\small
\setlength{\tabcolsep}{5pt}
\begin{tabularx}{\textwidth}{C{0.14\textwidth} C{0.14\textwidth} C{0.14\textwidth} Y Y Y}
\toprule
Cropping & Flipping & Relighting & Train & Validation & Test \\
\midrule
No & No & No & 1883 & 270 & 539 \\
Yes & No & No & 9415 & 1350 & 2695 \\
Yes & Yes & No & 9415 & 1350 & 2695 \\
Yes & No & Yes & 9415 & 1350 & 2695 \\
Yes & Yes & Yes & 9415 & 1350 & 2695 \\
\bottomrule
\end{tabularx}
\end{table}

Overall, these augmentation strategies simulate three main identified variations in this study. By cropping, spatial shifts in both training and test data can be achieved. Training data quantity can be augmented alongside. Horizontal/vertical flipping is able to play a similar role in introducing additional spatial shift. Relighting can be achieved by adjusting intensity values in each image, maintaining the absolute difference between raveling and non-raveling areas, and keeping the original classification label. This benchmarking design avoids the overhead of manual labeling, and because all transformations are derived from the original labeled dataset, the augmented data inherits the ground truth labels with high confidence.

\subsubsection{Design for Benchmarking Experiments}

Combining such variations, controlled experiments can be performed on any ML/DL models in multiple ways. In each experiment, specific variations can be injected into the training/validation and test data conveniently, with other variation factors held constant. As shown in Figure~\ref{fig:benchmark-heatmaps}, variations can be applied to training/validation data and test data separately. The vertical axis denotes the variations applied to the training data, while the horizontal axis represents the variations applied to the test data.

\textbf{Experiment with spatial shift.} We use cropped images from different spatial regions of the original data as described above. In the first three experimental rows in Figure~\ref{fig:benchmark-heatmaps}, the training data consists of cropped patches from the top-left corner of each original image, maintaining a fixed spatial perspective. In the next three rows, training images are randomly sampled from all corners and the center of the original image. This setup ensures spatial shift is included for comparison in training and enables researchers to assess the impact of spatial diversity on model performance. The last four rows further expand the spatial diversity by using all cropped regions in training, combined with horizontal/vertical flipping to simulate additional spatial variation. All trained models are evaluated on a consistently augmented test set using the same cropping procedure (denoted as “all” on the horizontal axis), allowing us to attribute differences in performance directly to the spatial characteristics of the training data.

\textbf{Experiment with training data quantity.} Within each spatial configuration, we vary the proportion of data used for training, while keeping the data distribution unchanged. In the first and second three-row groups, the training set includes 5\%, 10\%, and 20\% of the available data. Due to the cropping strategy, 20\% in these cases represents the full set of images available from one type of crop. In contrast, the last four rows utilize 100\% of the available cropped data, reflecting the full dataset. By holding other conditions constant, this setup enables us to isolate and quantify the contribution of data quantity to model performance.

\textbf{Experiment with difference in illumination scale.} In the last four training configurations, the first two rows use the full dataset without any lighting augmentation, while the last two rows include controlled intensity shifts (“relight” on the horizontal axis). All models are tested on a version of the test set that includes similar brightness perturbations. By comparing the performance between models trained with and without lighting variation, we can evaluate how well each approach generalizes to unseen illumination conditions.

\textbf{Experiment with combined variations.} Finally, our experimental design also supports studies on combined variations. This includes scenarios where multiple augmentations are applied simultaneously during training, as well as cases where researchers assess cross-variation effects on trained models. For example, in the former case, flipping and relighting can be applied to data together as shown in the last row. The latter refers to situations where a model trained with one type of variation, such as spatial shift, is tested on data exhibiting a different type of variation, such as lighting change. By analyzing performance across combinations of training and test conditions, we can explore whether training with certain types of variation improves robustness to others. The overall compound operation can be expressed in Equation~\eqref{eq:compound-augmentation}.

\begin{equation}
I_{\text{aug}} =
f_{\text{flip}}^{1_{\text{flip}}}
f_{\text{crop}}^{1_{\text{crop}}}
f_{\text{relight}}^{1_{\text{relight}}}(I)
\label{eq:compound-augmentation}
\end{equation}

\noindent where $1_{\text{operation}} \in \{0,1\}$; when $1_{\text{operation}} = 1$, the specific operation is applied. $1_{\text{crop}} = 1$ is always true during the benchmarking experiments, meaning cropping is constantly applied to all original images

Altogether, this benchmarking framework setup provides a flexible and scalable platform for evaluating models under realistic and diverse data conditions. The following sections present the results benchmarked with this framework and discusses their implications for model generalizability and implementation.

\section{Experiments}

\subsection{Experimental Setup}

\subsubsection{Data Preparation}

The original dataset is obtained from the previous study \cite{tsai2022review} and the methodology outlined in the preceding section is applied to it to construct our benchmarking dataset. The original data was collected by the Florida Department of Transportation (FDOT) using a pair of vehicle-mounted Laser Crack Measurement System (LCMS) sensors \cite{laurent2012}. These sensors employ line lasers and high-speed cameras to capture high-resolution 3D pavement surface data at highway speeds (up to 100 km/h). The sensor pair is mounted at the rear of the survey vehicle, enabling full-lane-width 3D data collection, as shown by Georgia Tech Sensing Vehicle (GTSV) in Figure~\ref{fig:data-collection} (a). The raw point clouds acquired by the sensors are processed into 3D pavement images, which performs both compression and rectification. The compression step rescales the height values of the point clouds into the range [0, 255], producing gray-scale range images. Rectification applies a Gaussian high-pass filter to eliminate gradual elevation changes caused by rutting or cross-slopes in the pavement surface.

In this study, each 3D pavement image has a resolution of 1,019 pixels (width) by 1,524 pixels (height). The distribution over four classes remains the same as \cite{tsai2022review} and quantity of the data under different augmentation settings is presented in Table~\ref{tab:data-quantity}, respectively. Cropping is applied to all images, resulting in a total data volume that is five times larger than the original dataset with the same data distribution and quality.

\subsubsection{Model Training}

As mentioned above, RF and ResNet are selected as baseline models for benchmarking due to their popularity and complementary strengths, with RF representing classical ML and ResNet representing DL approaches.

The Random Forest model is trained using handcrafted macrotexture features extracted from the input data \cite{hsieh2021,tsai2022review,tsai2015}. In our case, each cropped and augmented 3D pavement image is transformed into a 606-dimensional feature vector, which serves as the input to RF model after normalization. Following previous studies, 20 decision trees are included in the RF model. During training, each tree learns to partition the feature space to minimize classification error. Training is conducted without a GPU or a validation set.

The 50-layered ResNet model is trained end-to-end on the normalized range images. The training process uses a batch size of 8 and is conducted for up to 100 epochs. To prevent overfitting and reduce unnecessary computation, early stopping is applied based on validation performance. The model is optimized using the Adam optimizer \cite{kingma2014} with a learning rate of 0.00008 and the cross-entropy loss. Training is conducted with a RTX 4070 GPU. The augmentation applied to the validation set is kept the same as the training set. The best validation model weights are selected for testing.

\subsubsection{Evaluation Metric}

Accuracy is the evaluation metric used in this study due to its simplicity, interpretability, and direct relevance to the problem at hand. As denoted in Equation~\eqref{eq:accuracy}, accuracy measures the proportion of correctly classified samples over the total number of samples, providing a clear and intuitive indication of overall model performance.

Given that our primary objective is to assess the robustness and generalizability of different models when exposed to various data variations, such as changes in illumination, spatial shifts, and data quantity, accuracy serves as a concise summary of each model's effectiveness across different experimental settings.

\begin{equation}
\mathrm{Accuracy} = \frac{N_{\mathrm{correct}}}{N_{\mathrm{total}}}
\label{eq:accuracy}
\end{equation}

\noindent where $N_{\mathrm{correct}}$ is the number of correct predictions, and $N_{\mathrm{total}}$ is the total number of samples.

\subsection{Benchmarking Results}

\begin{figure}[H]
\centering
\includegraphics[width=0.78\textwidth]{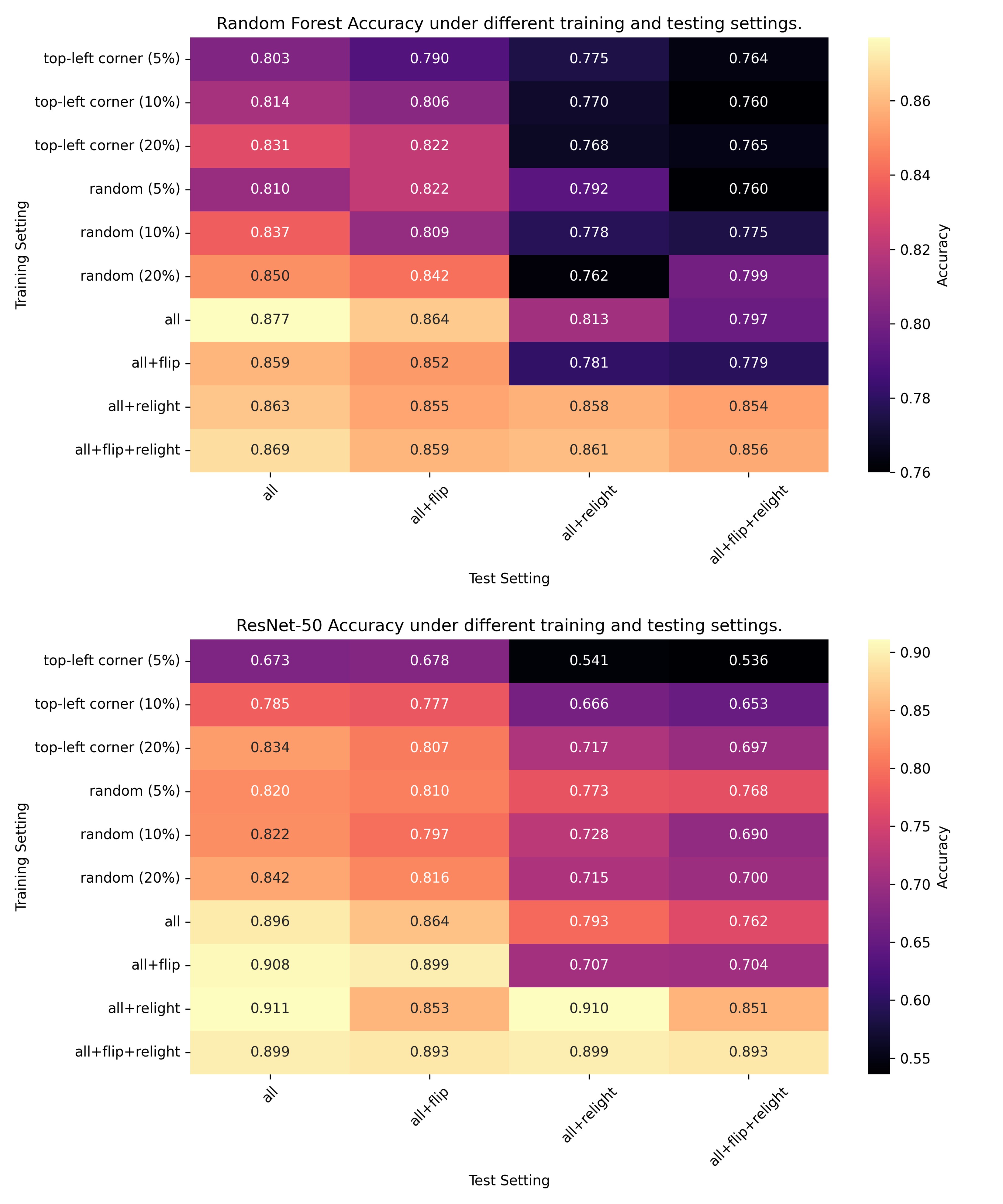}
\caption{Heatmaps of the benchmarking accuracy for Random Forest and ResNet-50 under different training and testing settings. These visual summaries correspond to the numerical results reported in Table~\ref{tab:rf-results} and Table~\ref{tab:resnet-results}.}
\label{fig:benchmark-heatmaps}
\end{figure}

We present the benchmarking results of the Random Forest and ResNet models in Table~\ref{tab:rf-results} and Table~\ref{tab:resnet-results}, respectively. These experiments are conducted based on the experimental design outlined in the previous section and illustrated in Figure~\ref{fig:benchmark-heatmaps}. The benchmarking framework systematically evaluates the impact of different types of variation in the data, including training data quantity, spatial and illumination diversity. Below, we summarize five key findings from the experiments.

\begin{table}[ht]
\caption{Random Forest accuracy under different training and testing settings. Under each test setting, the best performance is highlighted in bold, and the second best is underlined.}
\label{tab:rf-results}
\centering
\small
\setlength{\tabcolsep}{4pt}
\begin{tabularx}{\textwidth}{L{0.28\textwidth} Y Y Y Y}
\toprule
Train/Test & all & all+flip & all+relight & all+flip+relight \\
\midrule
top-left corner (5\%) & 0.803 & 0.790 & 0.775 & 0.764 \\
top-left corner (10\%) & 0.814 & 0.806 & 0.770 & 0.760 \\
top-left corner (20\%) & 0.831 & 0.822 & 0.768 & 0.756 \\
random (5\%) & 0.810 & 0.822 & 0.792 & 0.760 \\
random (10\%) & 0.837 & 0.809 & 0.778 & 0.775 \\
random (20\%) & 0.850 & 0.842 & 0.762 & 0.799 \\
all & 0.877 & 0.864 & 0.813 & 0.797 \\
all+flip & 0.859 & 0.852 & 0.781 & 0.779 \\
all+relight & 0.863 & 0.855 & 0.858 & 0.854 \\
all+flip+relight & 0.869 & 0.859 & 0.861 & 0.856 \\
\bottomrule
\end{tabularx}
\end{table}

\begin{table}[ht]
\caption{ResNet-50 accuracy under different training and testing settings. Under each test setting, the best performance is highlighted in bold, and the second best is underlined.}
\label{tab:resnet-results}
\centering
\small
\setlength{\tabcolsep}{4pt}
\begin{tabularx}{\textwidth}{L{0.28\textwidth} Y Y Y Y}
\toprule
Train/Test & all & all+flip & all+relight & all+flip+relight \\
\midrule
top-left corner (5\%) & 0.673 & 0.678 & 0.541 & 0.536 \\
top-left corner (10\%) & 0.785 & 0.777 & 0.666 & 0.653 \\
top-left corner (20\%) & 0.834 & 0.807 & 0.717 & 0.697 \\
random (5\%) & 0.820 & 0.810 & 0.773 & 0.768 \\
random (10\%) & 0.822 & 0.797 & 0.728 & 0.690 \\
random (20\%) & 0.842 & 0.816 & 0.715 & 0.700 \\
all & 0.896 & 0.864 & 0.793 & 0.762 \\
all+flip & 0.908 & 0.899 & 0.707 & 0.704 \\
all+relight & 0.911 & 0.853 & 0.910 & 0.851 \\
all+flip+relight & 0.899 & 0.893 & 0.899 & 0.893 \\
\bottomrule
\end{tabularx}
\end{table}

\textbf{Training data quantity.} Across all experimental settings, we observe that increasing the amount of training data with the same data distribution and quality consistently improves model performance. It is evident when the test conditions are held constant (in the same column), models trained with a higher proportion of data generally achieve better accuracy. For example, in the first column of Table~\ref{tab:rf-results} and Table~\ref{tab:resnet-results}, the accuracy of the same model constantly improves as more training data are used from 5\%, 10\%, 20\% and all. This reinforces a well-established understanding in ML: high-quality data volume positively correlates with model's performance. This effect holds true for both RF and ResNet in raveling detection, although the degree of improvement varies between the two.

\textbf{Spatial diversity in training.} When the quantity of training data is held constant, spatial diversity plays a crucial role when benchmarked against variations like spatial shift in test set. Models trained on data randomly sampled from various spatial locations within the cropped images outperform those trained on data fixed to a specific region, such as the top-left corner, as shown in the intersection between the first six rows and first two columns in Table~\ref{tab:rf-results} and Table~\ref{tab:resnet-results}. This suggests that exposure to a broader range of spatial contexts during training improves the model's robustness to spatial shifts that may occur in real-world implementation.

\textbf{Illumination variability in training.} Lighting is found to be a particularly sensitive factor. Even minor adjustments in brightness, such as adding or subtracting a small constant to pixel values in our approach, can cause noticeable degradation in model performance if the model was not trained with similarly relighted samples, as illustrated by the last two columns in both tables. This effect is more severe for ResNet than for RF, indicating that DL models are possibly more susceptible to illumination changes in this task. For instance, on the 7th row of each table, the accuracy of RF drops from 0.877 to 0.813 when the lighting variability is introduced in the test data, while for ResNet, the drop is more notable, from 0.896 to 0.793. Among all variations tested, changes in illumination lead to the most substantial drops in accuracy.

\textbf{Cross-variation robustness.} Training on data that includes one type of variation often helps the model become more robust to other similar types as well. For instance, models trained on randomly sampled spatial data also perform better under flipping transformations, even though flipping was not explicitly included in training. In both tables, the comparison between the 1st and 4th rows under the first two columns can be a good example. This suggests that introducing randomness and variability in training helps more robust feature representations and better generalization to unseen conditions.

\textbf{Model Sensitivity.} RF exhibits less sensitivity to data variations compared to ResNet. When exposed to new variations not seen during training, RF exhibits relatively less unstable performance. This is likely because RF relies on simpler decision boundaries based on discrete features, which have been carefully chosen by experts. ResNet learns representations during training and selects the best weights based on validation. However, since the validation set is derived from the training data and lacks unseen variations, the selected weights remain biased. As a result, RF is more resilient to perturbations in lighting or spatial composition in the test data. However, with larger and more diverse training data, ResNet consistently outperforms RF, demonstrating its capacity to leverage higher learning capacity when adequately trained.

The results from our benchmarking experiments indicate that models trained without sufficient variation in the training data are likely to struggle when evaluated on data containing such variations. It confirms that both data quantity and diversity in training, such as spatial, illumination, etc., are essential to building robust and generalizable models. For example, models achieve at least a 9.2\% gain in accuracy under the most diverse conditions (from 0.764 to 0.856 in the last column in Table~\ref{tab:rf-results}). Without them, seemingly minor differences in test data can lead to substantial performance degradation. Among the three types of variation evaluated, illumination difference has the most significant negative impact. A case study discussed in the following section illustrates an example of the difference brought by illumination discrepancies.

\section{A Case Study In Implementation}

A comparative case study is conducted using three years of data (2014-2016) collected along a segment of I-59 (mileposts 12 to 14) in Georgia, U.S. The data was acquired by the GTSV, following the same collection procedure as in Figure~\ref{fig:data-collection} (a). Each year of data includes approximately 647 images. The DL model ResNet is employed in this case study due to its unified feature extraction and classification pipeline, allowing efficient validation of hypotheses without the need for separate feature extraction. Notably, the dataset is not paired with ground truth annotations, similar to real-world scenarios in implementation. Consequently, year-to-year consistency is employed for evaluating the model's performance. A fundamental rule is that the raveling severity of the current year cannot be lower than last year or higher than next year either.

\begin{figure}[H]
\centering
\includegraphics[width=\textwidth]{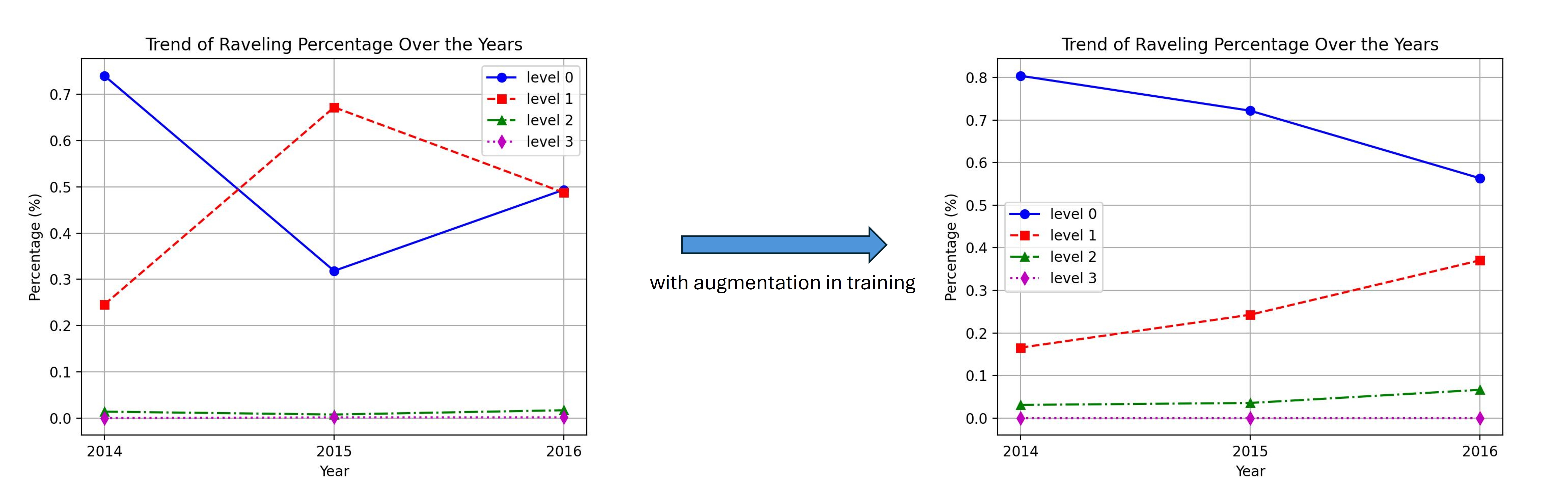}
\caption{Raveling severity level distribution across years based on prediction results. The left panel shows results from the model trained with the original dataset without data augmentation, while the right panel shows results from the model trained with data augmentation. In this case, the data augmentation introduces more lighting differences into the training data.}
\label{fig:year-distribution}
\end{figure}
\begin{figure}[H]
\centering
\includegraphics[width=\textwidth]{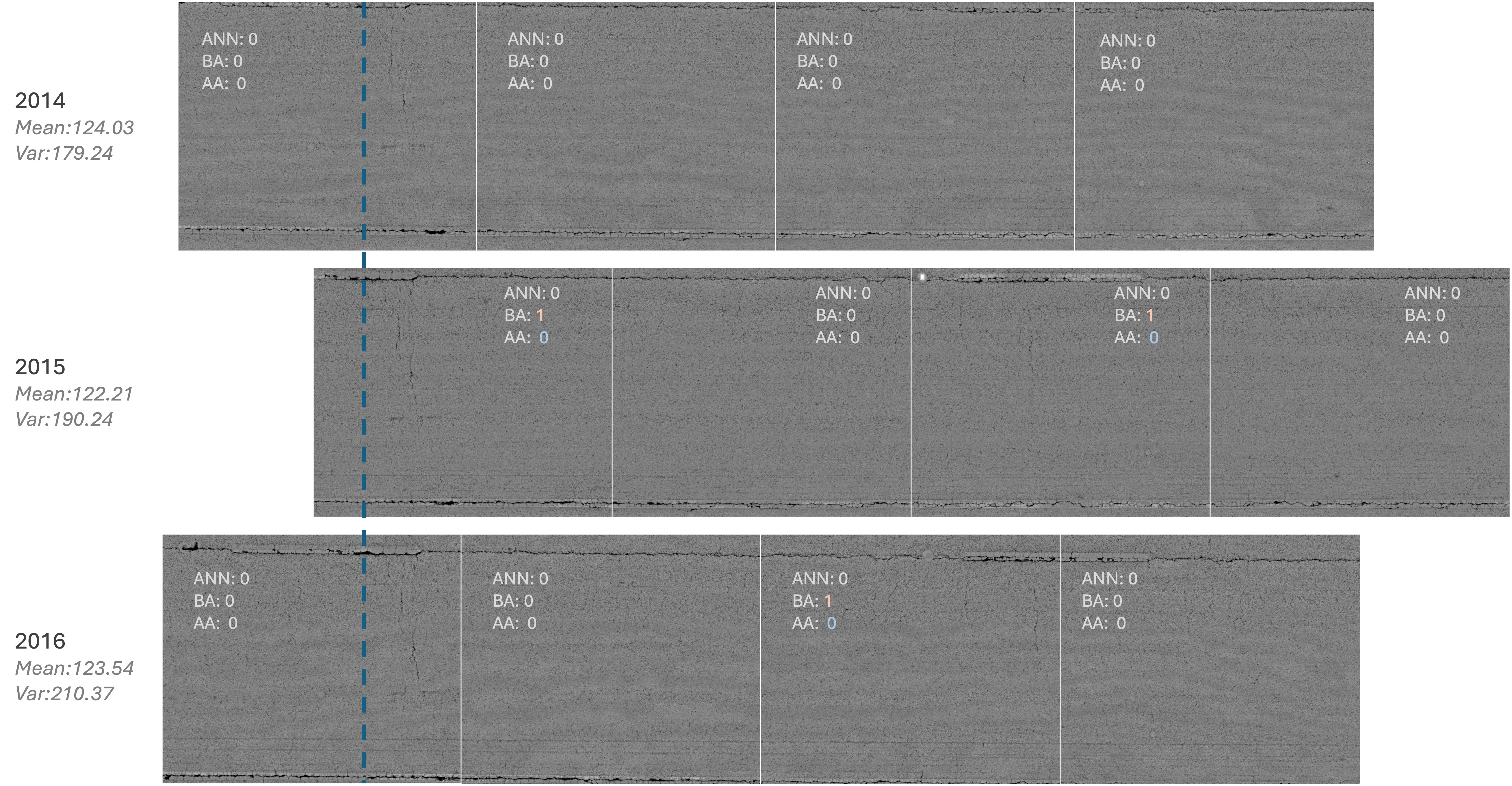}
\caption{Examples of spatially aligned multi-year data from 2014, 2015, and 2016. The mean and variance values of each year's data are displayed on the left. The blue dashed line marks the referenced visual features for image alignment. ANN, BA, and AA stand for manual annotation, before augmentation prediction, and after augmentation prediction.}
\label{fig:multi-year-examples}
\end{figure}

\textbf{Results without data augmentation in training.} The ResNet model trained on the original dataset from \cite{tsai2022review} is first applied to the multi-year data. Results reveal inconsistencies in prediction outcomes across different years. As shown in the left panel of Figure~\ref{fig:year-distribution}, the proportion of severity level 0 decreases in 2015 but increases again in 2016, which is an impossible pattern given the fact that no maintenance activities occurred between these two years.

Some examples are visualized in Figure~\ref{fig:multi-year-examples}, confirming misclassifications, such as images without visible raveling (L0) being predicted as low raveling (L1). To investigate this phenomenon, the study calculates the mean and variance of the image data of each year, labeled in Figure~\ref{fig:multi-year-examples}. It is found that the mean pixel intensity of the 2015 dataset decreases from 124.03 in 2014 to 122.21, along with an increase in variance. Although these changes are subtle and largely negligible to human observers, we hypothesize that they are the main factor causing the inconsistency based on benchmarking findings and can be categorized as lighting variation because of the direct change in pixel intensity.

\textbf{Results with data augmentation in training.} To confirm and mitigate this issue, the model is retrained using an augmented dataset that incorporates lighting diversity. Departing from the fixed perturbations used in the benchmark framework, this approach applies random adjustments to the mean (±0.03) and variance (±0.01) of the normalized training images to better fit the random nature of real-world data. The resulting model exhibits improved consistency across the three annual data. This enhancement is evident both in the corrected predictions for previously inconsistent samples in Figure~\ref{fig:multi-year-examples} and in the more coherent trends shown in the right panel of Figure~\ref{fig:year-distribution}. Consistency analysis plays an important role since no ground truth is usually available in large-scale implementation.

This case study illustrates that training with variation-diverse data (illumination in this case), aligned with benchmarking findings, enhances the robustness and temporal consistency of model predictions. This multi-year setting and improvement can further support annual trend analysis, which is valuable for long-term pavement maintenance planning. Future work can focus on designing evaluation metrics that more explicitly measure consistency in predictions across years.

\section{Conclusions And Recommendations}

This work studies the problem of performance degradation observed when well-trained ML/DL models are applied to real-world data for raveling detection. It begins by hypothesizing that variations in the data are the primary cause of this issue based on practical experience and data observation. To investigate this, a benchmarking framework (RavelingArena) is developed to evaluate the performance of models systematically and quantitatively for raveling detection under identified common forms of data variation, including changes in training data quantity, illumination difference and spatial shift. Experimental results show support for the initial hypothesis, demonstrating that both ML and DL models benefit substantially from increased training data quantity and diversity, maintaining consistent high performance in classification accuracy (around 86\% for RF and 90\% for ResNet-50) across varied test conditions. These findings emphasize the importance of incorporating real-world variability into model training, particularly for large-scale implementation.

Building on these insights, a multi-year case study is conducted using unannotated field data collected along the same road segment across three consecutive years. The case study initially uses a DL model trained on the original dataset and reveals inconsistencies in predictions, largely attributed to subtle lighting variations by both consistency analysis. Retraining the model with illumination-augmented data results in more consistent predictions across years, as revealed by benchmarking experiments. In the absence of ground truth annotations, a common scenario in real-world deployments, year-to-year consistency serves as an indirect but practical indicator of improved model performance. These results highlight the practical value of benchmarking experiments and confirm the applicability of their findings to real-world implementation.

The proposed benchmarking approach provides a scalable and effective tool for guiding the design and validation of robust pavement condition assessment models in implementation settings. For future research, the following items are recommended:

\begin{itemize}
\item Develop formal metrics to quantify prediction consistency across temporal datasets, with support for automated spatial alignment \cite{yang2025,zitova2003}.
\item Investigate additional sources of variation, such as sensor differences, changes in pavement material properties, variation in data distribution, etc. \cite{zhang2023}.
\item Integrating adaptive augmentation techniques into training pipelines to improve model robustness \cite{shorten2019}, along with further exploration into resilient model architectures and task formulations designed to handle data variability more effectively \cite{yu2022,zhang2019}.
\end{itemize}

\section{Acknowledgments}

The authors would like to thank the Georgia Department of Transportation (GDOT), Florida Department of Transportation (FDOT), Mississippi Department of Transportation (MDOT), and Dr. Ryan Salameh for their valuable input and discussion. Grammarly was used to check grammar in this paper.

\section{Author Contributions}

The authors confirm their contribution to the paper as follows: Study conception and design: X. Zhang, H. Wang, Z. Yang, and Y. Tsai; Methodology development: X. Zhang; Data collection coordination: Y. Tsai; Experiments and analysis: X. Zhang; Draft manuscript preparation: X. Zhang; Manuscript review and editing: X. Zhang, H. Wang, Z. Yang, and Y. Tsai.

\section{Declaration Of Conflicting Interests}

The authors declared no potential conflicts of interest with respect to the research, authorship, and/or publication of this article.



\FloatBarrier
\bibliographystyle{unsrtnat}
\bibliography{references}

\end{document}